\documentclass{article}

\usepackage{PRIMEarxiv}

\usepackage[utf8]{inputenc} % allow utf-8 input
\usepackage[T1]{fontenc}    % use 8-bit T1 fonts
\usepackage{hyperref}       % hyperlinks
\usepackage{url}            % simple URL typesetting
\usepackage{booktabs}       % professional-quality tables
\usepackage{amsfonts}       % blackboard math symbols
\usepackage{nicefrac}       % compact symbols for 1/2, etc.
\usepackage{microtype}      % microtypography
\usepackage{lipsum}
\usepackage{graphicx}
\graphicspath{{media/}}     % organize your images and other figures under media/ folder

\usepackage{algorithm}
\usepackage{algorithmic}

\usepackage{microtype}
\usepackage{
graphicx}
\usepackage{subfigure}
\usepackage{booktabs} % for professional tables

\usepackage{hyperref}

\usepackage{amsmath}
\usepackage{amssymb}

\usepackage{mathtools}
\usepackage{amsthm}

\usepackage[capitalize,noabbrev]{cleveref}

\theoremstyle{plain}

\theoremstyle{definition}

\theoremstyle{remark}
\newtheorem{
remark}[theorem]{Remark}

\usepackage[textsize=tiny]{todonotes}
  
\title{Neural Control: Concurrent system identification and control learning with neural ODE}

\author{
  Cheng Chi \\
  University of Toronto \\
  \texttt{cheng.chi@mail.utoronto.ca} \\
}

\begin{document}
\maketitle

\begin{abstract}
Controlling continuous-time dynamical systems is generally a two step process: first, identify or model the system dynamics with differential equations, then, minimize the control objectives to achieve optimal control function and optimal state trajectories. However, any inaccuracy in dynamics modeling will lead to sub-optimality in the resulting control function. To address this, we propose a neural ODE based method for controlling unknown dynamical systems, denoted as Neural Control (NC), which combines dynamics identification and optimal control learning using a coupled neural ODE. Through an intriguing interplay between the two neural networks in coupled neural ODE structure, our model concurrently learns system dynamics as well as optimal controls that guides towards target states. Our experiments demonstrate the effectiveness of our model for learning optimal control of unknown dynamical systems. The relevant code implementations can be accessed at \href{https://github.com/chichengmessi/neural_ode_control/tree/main}{this GitHub link}. 
\end{abstract}

 % Another set of approaches - reinforcement learning, when applied for continuous-time optimal control problems, folds the dynamics modeling into controller training via value function approximation or policy gradient through extensively interacting with the environment, but it suffers from low data efficiency. 

% keywords can be removed

\section{Introduction}
\label{intro}

Optimal control is a robust mathematical framework that enables finding control and state trajectories for dynamical systems over a period of time to minimize a performance metric \cite{oc}. The primary objective in optimal control is to determine a control law that minimizes or maximizes a performance metric, such as energy consumption, time to reach a target, or cost of operation. Solving optimal control problems can be difficult, especially for systems with nonlinear dynamics. Classical optimal control theory, a cornerstone of modern control systems engineering, has evolved through seminal contributions that establish principles for optimizing dynamic system behaviors. Central to this evolution is the calculus of variations \cite{gelfand1963calculus}, providing foundational techniques for formulating and solving optimization problems in continuous systems. Another pivotal contribution is Pontryagin's Maximum Principle \cite{max_principle1} \cite{max_principle2} offering necessary conditions for optimality in control problems, widely applicable in fields from aerospace to economics. The Hamilton-Jacobi-Bellman (HJB) equation, foundational in dynamic programming, extends these concepts to more complex systems \cite{bellman1957dynamic}. These works collectively underpin the theoretical framework of classical optimal control. 

While classical approaches to optimal control are powerful, they often hinge on the analytical resolution of complex equations, a task that can be daunting and computationally intensive. Recent advancements have seen the integration of Machine Learning methods, particularly artificial neural networks (ANNs), into the domain of optimal control. Works such as those by authors in \cite{AI_Pontryagin} and \cite{first_NODEC} demonstrate the potential of embedding parameterized control functions within a system's dynamical equations. These approaches leverage auto-differentiation capabilities of ANNs to iteratively refine and optimize control functions, a significant leap from traditional methods.

However, a critical assumption underlying these AI-enhanced methods is the prior knowledge of system dynamics. Typically, this knowledge is encapsulated in differential equations, formulated through system identification – a process of building mathematical models from empirical data. Traditionally, system identification relies on statistical methods as elucidated in \cite{system_identification_soderstrom1989system}. The advent of modern machine learning techniques has also revolutionized this process, enhancing both efficiency and accuracy in modeling dynamic systems. Pioneering works in this area, such as those by \cite{chen2022neural}, \cite{author2023accurate}, and \cite{holl2020learning}, exemplify the successful application of these advanced techniques in system identification tasks.

Building on these developments, we introduce - Neural Control (NC), which synergizes system identification with the learning of optimal control, specifically tailored for continuous time dynamical systems. This framework eliminates the need for separate system identification prior to control optimization in optimal control framework. By employing a coupled neural ordinary differential equations, NC learns the underlying system dynamics while simultaneously learning the optimal control function for continuous time optimal control. 

Progress in automatic differentiation and machine learning methodologies has significantly enhanced the evolution of modeling and control strategies in the domain of optimal control. For instance, reinforcement learning is an algorithmic framework that can be applied for optimal control problems - generally in an unknown environment. Even though certain RL algorithms can be used to approximate the solution to the Hamilton-Jacobi-Bellman (HJB) equation (necessary and sufficient condition for optimality), recent advancements in RL such as the policy gradient-based methods \cite{TRPO} \cite{PG_RL} or the actor-critic based methods \cite{AC_RL} are in a purely data-driven regime. They directly parameterize the control policy and collect a large number of interactions with the environment to build differentiable loss for improving the parameterized policy, thus skipping the dynamics modelling and the derivation of optimality condition as in the principle-driven approach. A subroutine in RL that involves dynamics modelling is known as the model-based RL, where the approximation of unknown dynamics amounts to a supervised learning problem, and the learned dynamics act as a subroutine in the original RL policy optimization process in the form of forward planning \cite{model-based-RL}. However, as dynamics modeling and policy learning are often separate processes, powerful policy optimizers exploit inaccuracies in learned dynamical models leading to adversarial examples and sub-optimal policies \cite{diffuser}. Besides, RL algorithms are known for having a low data efficiency \cite{low_eff_RL}, and the available data can be extremely limited for some dynamical systems \cite{rim}. 

Neural Ordinary Differential Equations (Neural ODEs) have demonstrated considerable efficacy as a framework for modeling continuous systems using time series data that is sampled irregularly \cite{neuralODE}. There has been a growing interest in the application of Neural ODEs within the realm of continuous-time control tasks. Notably, the use of Neural ODEs has been explored in the domain of learning optimal control functions, as evidenced in studies such as \cite{AI_Pontryagin}, \cite{first_NODEC}, \cite{Boettcher2022}, and \cite{neural_odes_feedback_2022}. These investigations have incorporated a parameterized controller within the actual dynamical functions $f$, revealing that such controllers can emulate optimal control signals by minimizing a loss function, circumventing the explicit resolution of maximum principle or Hamilton-Jacobi-Bellman (HJB) equations.

Despite these advancements, existing methodologies predominantly focus on controlling continuous-time dynamical systems when the dynamical functions are known. Conversely, the inherent suitability of Neural ODEs for modeling continuous dynamical systems has prompted their extensive application in learning the dynamics of the underlying system or system identification tasks, as highlighted in \cite{chen2022neural}, \cite{learning_dynamics_partial_2022}, and \cite{differential_neural_network_2023}.

However, to the best of our knowledge, there has been no methodological integration of system identification and the learning of optimal control policies within a singular Neural ODE framework. In response to this gap, we introduce Neural Control (NC). NC represents a novel approach that combines the dual processes of learning the optimal control function and identifying system dynamics. This integration is executed within a coupled Neural ODE structure, enabling concurrent optimal control function learning and system identification in an end-to-end manner. This approach aims to harness the full potential of the Neural ODE model for controlling continuous-time systems.

\section{Method}

In this section, we introduce our method: Neural Control (NC), a coupled neural ODE structure that simultaneously learns optimal control as well as system dynamics of unknown dynamical systems. We first briefly introduce few relevant concepts before discussions of NC. 

% \begin{algorithm}
% \caption{Quantum Neural ODE Training Algorithm}
% \begin{algorithmic}[1]
% \REQUIRE Initial neural network weights for neural ODE
% \REQUIRE Initial state: $\psi_0$
% \REQUIRE Parameterized circuit: $U(\theta_t)$
% \REQUIRE Initial circuit parameters: $\theta_0$
% \REQUIRE Encoding basis states: \textit{encoding basis states}

% \FOR{$\text{epoch} = 1$ to $n_{\text{epochs}}$}

%     \STATE $\text{time\_batch} \gets \text{Randomly sample times from [0, T]}$
%     \STATE $\text{thetas at different t} \gets \text{neural\_ODE}(\theta_0, \text{time\_batch})$
%     \STATE $\text{loss} \gets 0$
    
%     \FORALL{$\theta_t$ in $\text{thetas at different t}$}
    
%         \STATE $U_t \gets U(\theta_t)$
%         \STATE $\psi_t \gets U_t \psi_0$
%         \STATE $\psi_{t\_sub} \gets \text{project\_to\_basis}(\psi_t, \text{encoding\_basis\_states})$
%         \STATE $\text{target\_values} \gets \text{get\_targets\_from\_MC\_or\_PDE()}$
%         \STATE $\text{loss} \gets \text{loss} + \text{MSE}(\psi_{t\_sub}, \text{target\_values})$
        
%     \ENDFOR
    
%     \STATE $\text{train neural ODE with adjoint backpropagation}(\text{loss})$
    
% \ENDFOR

% \end{algorithmic}
% \end{algorithm}

\subsection{Classical optimal control}
In this section, we briefly explain the classical methods for solving optimal control problems and particularly the Maximum principle \cite{max_principle1}.

The first implicit step of controlling an unknown system would be to try to formulate system dynamics as a system of ODEs, which relies on prior knowledge or requires fitting the parameters in the ODEs. Once the dynamics are estimated and represented as $f(x,u,t)$, where $x$ represents the state variable, $u$ is the control input, and $t$ denotes time, we can express the system as follows:
\begin{equation}
\begin{aligned}
\label{control_dyn}
\dot{x}(t) = f(x(t), u(x,t), t) \\
\end{aligned}
\end{equation}
Then we formulate our control objective as minimization of the functions shown in equation (\ref{control_obj}):
\begin{equation}
\label{control_obj}
J(u) = \int_{0}^{T} L(x(t), u(x,t), t) \, dt + M(x(T))
\end{equation}
where $M(x(T))$ specifies the terminal cost that penalizes the state attained at T, and $L(x(t), u(x, t), t)$ represents the instantaneous cost. The subsequent steps involve converting the constrained optimization problem (equation (\ref{control_dyn}) as constraints) into unconstrained optimization via Lagrange multiplier, redefining terms inside that functional with the quantity known as Hamiltonian, and applying a first-order necessary optimality condition on the augmented objective functional. Eventually we reach the necessary condition for the optimality of the control: if $(u^*, x^*)$ is optimal, then:
\begin{equation}
\begin{aligned}
\label{eq:max_principle}
\dot x_i = \frac{\partial H}{\partial \lambda_i}  \quad \dot \lambda_i = -\frac{\partial H}{\partial x_i} \quad \forall i \\
\lambda(T) = \frac{\partial M}{\partial x} (x(T)) \\
H(x^*(t),u^*(x,t),\lambda^*(t)) \leq  H(x^*(t),u,\lambda^*(t)) \quad \forall u 
\end{aligned}
\end{equation}
where $\lambda^*(t)$ are the costate variables associated with $x^*(t)$. The aforementioned optimality condition typically leads to a system of differentiable equations, and the solutions to this system contain the desired optimal control. This principle-driven approach is often analytically and computationally intractable when applied to complex dynamical systems. 

\subsection{Neural ODE} 
Neural ODE (Ordinary Differential Equation) \cite{neuralODE} is a powerful modelling framework that leverages differential equations and neural networks to learn continuous-time dynamics. The key equation of Neural ODE is:
\begin{equation}
\setlength{\abovedisplayskip}{0.1pt}
\setlength{\belowdisplayskip}{0.1pt}
\begin{aligned}
\frac{{dh(t)}}{{dt}} = f_{\theta}(h(t), t)
\end{aligned}
\end{equation}
where \(h(t)\) represents the hidden state at time \(t\), \(f\) is a neural network parameterized by \(\theta\), and \(t\) denotes the continuous time variable. The initial condition for Neural ODE is given by $h(t_0) = h_0$. The solution of the Neural ODE or the model predictions can be obtained by numerically integrating the above equation over the desired time interval. The efficient back-propagation for training neural ODE is achieved by the adjoint method. It requires solving a backward-in-time second ODE, known as the adjoint ODE, which is associated with the original ODE. By solving the adjoint ODE, we obtain gradients of the loss function with respect to the state variables of the original ODE at each time step. These gradients are then utilized, along with the chain rule, to compute gradients with respect to the parameters of the neural network \cite{nerualODE}.

\subsection{Neural Control: NC}
\subsubsection{Model architecture}
Our NC model consists of a coupled neural ODE structure. We still parameterize the control function $u(x,t)$ through a neural network $  h_{\theta}(x, t)$, denoted as the controller. This controller is then embedded as part of the dynamics represented by another neural network $  g_{\gamma}(x,u,t)$ denoted as the dynamics learner, and the control signal $u$ is given by the controller. This is represented by the following equations manifested by a coupled neural ODE structure: 
\begin{equation}
\begin{aligned}
\label{NODEC}
\dot{x} =  {g}_{\gamma}(x, {h}_{\theta}(x, t), t) \\ 
\end{aligned}
\end{equation}
This is essentially a coupled neural ODE where the neural part consists of two neural networks: controller $  h_{\theta}$ and dynamics learner $  g_{\gamma}$. The integration in Equation (\ref{NODEC}) is conducted using a numerical solver over the dynamics specified by the coupled $  h_{\theta}$ and $  g_{\gamma}$, and this is how NC makes a prediction about the state $x_T$ as a result of continuously controlling the state evolution using the controller in the environment with unknown dynamics. For instance, starting at $x_0$, NC predicts where the system will end up by: 
\begin{equation}
\begin{aligned}
\label{NODEC_integral}
x_{T} = x_{0} +  \int_{0}^{T}  {g}_{\gamma}(x_{t}, {h}_{\theta}(x_{t}, t), t)dt 
\end{aligned}
\end{equation}

\subsubsection{Training Loss}
There are two objectives when we train our NC model given by $ {g}_{\gamma}(x, {h}_{\theta}(x, t), t)$. The first objective is for the dynamics learner $\hat{g}_{\gamma}$ to approximate the real unknown dynamics $f(x,u,t)$: the integration in Equation (\ref{NODEC}) gives us the model prediction of the trajectory roll-out, and we want this prediction to match the real trajectory coming from the environment (the real dynamics) using the same controller $ {h}_{\theta}$. The second objective is for the controller ${h}_{\theta}$: we want to satisfy the optimal control objectives, for example, by continuously applying control signal u given by $ {h}_{\theta}$, the end state in the predicted trajectory $x_T$ using the dynamics learner $ {g}_{\gamma}$ (which is also the end state in the real trajectory if ${g}_{\gamma}$ is well trained) should be close to our target state $x^*$. 

These two objectives manifest the training loss, which is presented in Figure \ref{fig:loss} below:
\begin{figure}[ht]
\vskip -0.1in
\begin{center}
\centerline{\includegraphics[width=\columnwidth]{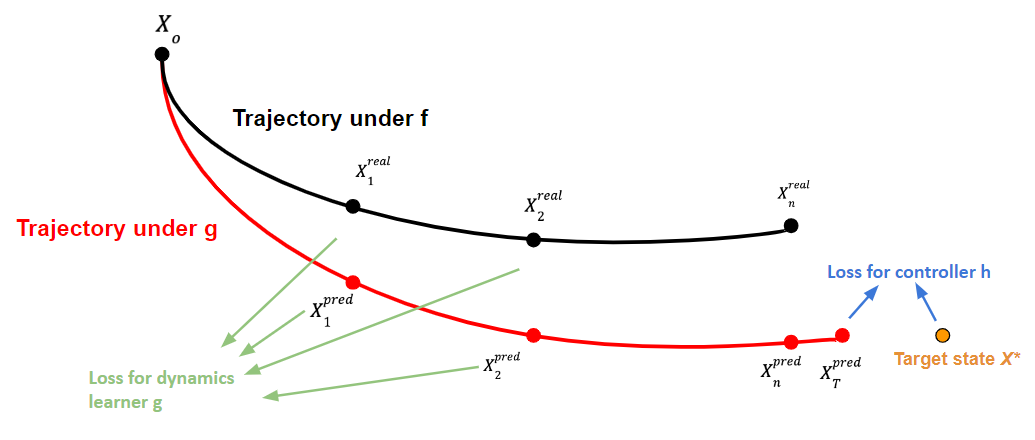}}
\caption{Training loss for dynamics learner $g_{\gamma}$ and controller $h_{\theta}$}
\label{fig:loss}
\end{center}
\vskip -0.2in
\end{figure} \\
Starting from the same initial state $x_0$, the predicted trajectory under $ {g}_{\gamma}(x, {h}_{\theta}(x, t), t)$ is shown in red, and the real trajectory coming from applying the same controller $ {h}_{\theta}$ in the environment or in real dynamics $f$ is shown in black. As $ {g}_{\gamma}$ should approximate $f$, the mean squared error (MSE) between these two trajectories (shown in green) is used to update $ {g}_{\gamma}$ to reflect the dynamic learning objective. In contrast, the MSE loss between $x_T^{pred}$ and $x^*$ (shown in blue) is used to train controller $ {h}_{\theta}$ to reflect the optimal control objective. Note that the controller loss can also be constructed using the whole trajectory, which is the case for our CartPole experiment in Section \ref{Exp}, and we use a single end state here for illustrative purpose. 

By setting up the losses in this manner, the training of the controller is conducted exclusively through simulation using roll-out by $ {g}_{\gamma}$ (as depicted by the blue arrows in Figure \ref{fig:loss}), and the real trajectories data from the environment is only required for training dynamics learner. This significantly improves data efficiency, which is demonstrated by our results in section \ref{Exp} where we successfully learn to control the CartPole system with less than 60 trajectories collected from the environment. This is extremely data-efficient compared to traditional RL methods that use hundreds or thousands of trajectories or episodes for CartPole \cite{cartpole_RL}.   

The relationship between controller ${h}_{\theta}$ and dynamics learner ${g}_{\gamma}$ is intriguing. On one hand, the dynamics learner determines whether the controller training is effective or not, as it is purely using trajectory roll-outs under $ {g}_{\gamma}$. If ${g}_{\gamma}$ has already been well-trained and sufficiently approximates $f$, the controller training process genuinely enhances the controller's performance in terms of our control objective: guiding states towards the target state under the real and unknown dynamics $f$. If the dynamics learner is poorly trained, the controller may exploit the learned dynamics and produce ineffective controls when deployed in the real dynamics. 

On the other hand, the controller plays a vital role in shaping the trajectory's roll-out within the state space, as its outputs serve as inputs to both the dynamics learner ${g}{\gamma}(x, {h}{\theta}(x, t), t)$ and the true dynamics $f(x, {h}_{\theta}(x, t))$. As indicated in Figure \ref{fig:loss}, these trajectories roll-outs establish the training loss for the dynamics learner. Therefore,  the controller essentially determines which portion of the state space dynamics the dynamics learner would fit. Through this mechanism, the controller effectively directs the dynamics learner's training process. 

\subsubsection{Alternative Training}
\label{sec:alternative training}
A naive approach for training NC model ${g}_{\gamma}(x, {h}_{\theta}(x, t), t)$ is a two-stage training process. In the first stage, the dynamics learner $ {g}_{\gamma}$ is trained to accurately capture the underlying dynamics $f$ by leveraging a diverse set of  controllers (ideally all possible control functions) for generation the input controls u. In the second stage, the controller $ {h}_{\theta}$ is trained within the learned dynamics model $ {g}_{\gamma}$ to attain optimal control, which is similar to the approach described in \cite{AI_Pontryagin} for learning optimal control with known dynamics.

There are two shortcomings with this two-stage training approach: First, once the training of the controller begins, no further updates will be made to the learned dynamics $ {g}_{\gamma}$. Consequently, the controller has the potential to exploit the fixed learned dynamics and generate adversarial controls if $ {g}_{\gamma}$ is poorly fit. This raises uncertainty regarding whether the set of controllers used in the initial stage of training $ {g}_{\gamma}$ is truly diverse enough for exploring all the state space of $f$ for a good fit of $ {g}_{\gamma}$. 

Furthermore, it is unnecessary to precisely fit $f$ over all state space. As mentioned earlier, the controller dictates which portion of the vector field the dynamics learner should approximate. Since our goal is to train the controller to be optimal, our focus is on thoroughly exploring the vector field of $f$ that is relevant to the optimal controller and controllers from previous training steps only. Therefore, we do not need to generate all possible control functions (and we may not even be able to generate near-optimal controllers from the start) for training the dynamics learner in the first stage. Instead, the dynamics learner only needs to approximate the vector field locally, specifically in the regions where the current controller is leading, which is shown in Figure \ref{fig:vector} Appendix \ref{app:motivation}. 
\begin{figure}[ht]
\begin{center}
\centerline{\includegraphics[width=\columnwidth]{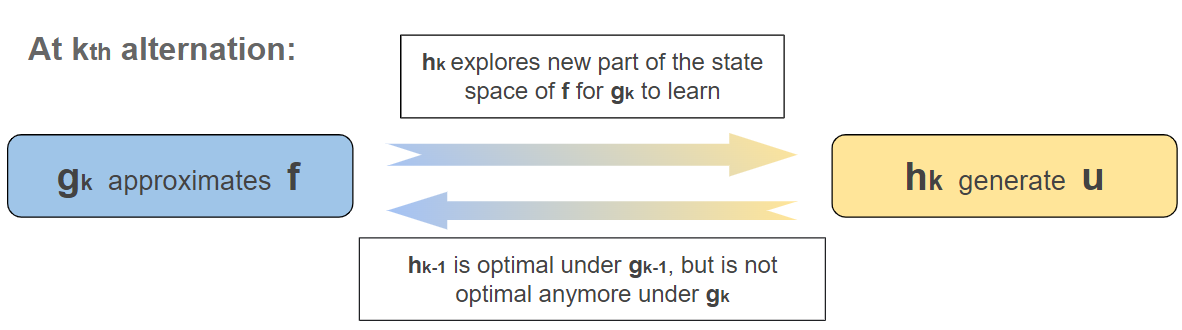}}
\caption{The $k_{th}$ alternation of the dynamics learner $ {g}_{\gamma}$ and controller $ {h}_{\theta}$ alternative training}
\label{fig:alternative train}
\end{center}
\vskip -0.3in
\end{figure} 

This naturally leads to the alternative training scheme depicted in Figure (\ref{fig:alternative train}) for training NC model $ {g}_{\gamma}(x, {h}_{\theta}(x, t), t)$. In this scheme, we alternate the training of the two neural networks, $ {g}_{\gamma}$ and $ {h}_{\theta}$, for in total of K iterations. We denote the neural networks at the $k_{th}$ alternation as $ {g}_{\gamma}^k$ and $ {h}_{\theta}^k$, respectively.

Within each alternation, such as the $k_{th}$ alternation, we first train the dynamics learner $ {g}_{\gamma}^k$ using numerous gradient steps to obtain $ {g}_{\gamma}^{k+1}$ while keeping the controller $ {g}_{\gamma}^k$ fixed. After this training, this new dynamics learner has learned to approximate the real dynamics  near the trajectories that the current controller $ {h}_{\theta}^k$ leads to.

Subsequently, we fix the dynamics learner and focus on training the controller $ {h}_{\theta}^k$ (only optimal under $ {g}_{\gamma}^{k}$) to be optimal under ${g}_{\gamma}^{k+1}$ in terms of the control objective, resulting in a new controller $ {h}_{\theta}^{k+1}$. However, this new controller would lead to a different portion of the state space whose dynamics may not be accurately approximated by the current dynamics learner ${g}_{\gamma}^{k+1}$, as ${g}_{\gamma}^{k+1}$ only approximates the true dynamics well near the trajectories associated with the previous controller $ {h}{\theta}^{k}$. Therefore, we proceed to the next iteration. This alternative process continues until convergence is achieved for both the dynamics learner and the controller. 

The algorithm for alternative training with the corresponding loss definitions is outlined in Algorithm \ref{alg:alternative_training}. It is important to note that, for the purpose of illustration, the algorithm is presented with only one initial state. However, it can be readily extended to handle a batch of initial states.
\begin{algorithm}[tb]
  \caption{Alternative Training of NC Model}
  \label{alg:alternative_training}
  \begin{algorithmic}[1]
   \STATE {\bfseries Input:} Initial state $x_0$, Target state $x^*$, Initialized neural networks $g_{\gamma}^{0}$ and $h_{\theta}^{0}$, Number of alternations $K$ between training $g_{\gamma}$ and $h_{\theta}$, Training iterations $n_g$ and $n_h$ within each alternation for training $g_{\gamma}$ and $h_{\theta}$ respectively, Interactive environment $f$

    \STATE Initialize $k \leftarrow 0$
    \WHILE{$k < K$}
      \STATE Deploy $ {h}_{\theta}^k$ in environment $f$ starting from $x_0$
      \STATE Collect real trajectories $T_{real}$
      \STATE \textbf{Train Dynamics Learner}
      \STATE For $n_g$ update steps:
      \STATE \quad Deploy $ {h}_{\theta}^k$ in $ {g}_{\gamma}^k$ starting from $X_0$
      \STATE \quad Collect predicted trajectories $T_{pred}$
      \STATE \quad Update $ {g}_{\gamma}^k$ with dynamics fitting loss: 
      \STATE \quad \quad \quad \quad $L_g = MSE(T_{pred}, T_{real})$
      \STATE \textbf{Train Controller}
      \STATE For $n_c$ update steps:
      \STATE \quad Deploy $ {h}_{\theta}^k$ in already updated $ {g}_{\gamma}^k$ starting from $X_0$
      \STATE \quad Collect predicted trajectories $T_{pred}$
      \STATE \quad Update $ {g}_{\gamma}^k$ with control loss: 
      \STATE \quad \quad \quad \quad $L_c = MSE(T_{pred}, x^*)$
      \STATE $k \leftarrow k + 1$
    \ENDWHILE
  \end{algorithmic}
\end{algorithm}

\section{Experiments}
\label{Exp}
\subsection{Control task: Linear System}
\label{sec:AxBu}
We now investigate the control performance of NC applied to a linear dynamical system described by $f(x,u)=Ax+Bu$. This is the same tasked used in \cite{AI_Pontryagin} and \cite{example1}.

\textbf{Dynamical system and control objectives}

\begin{figure*}[!t]
\centerline{\includegraphics[width=\textwidth]{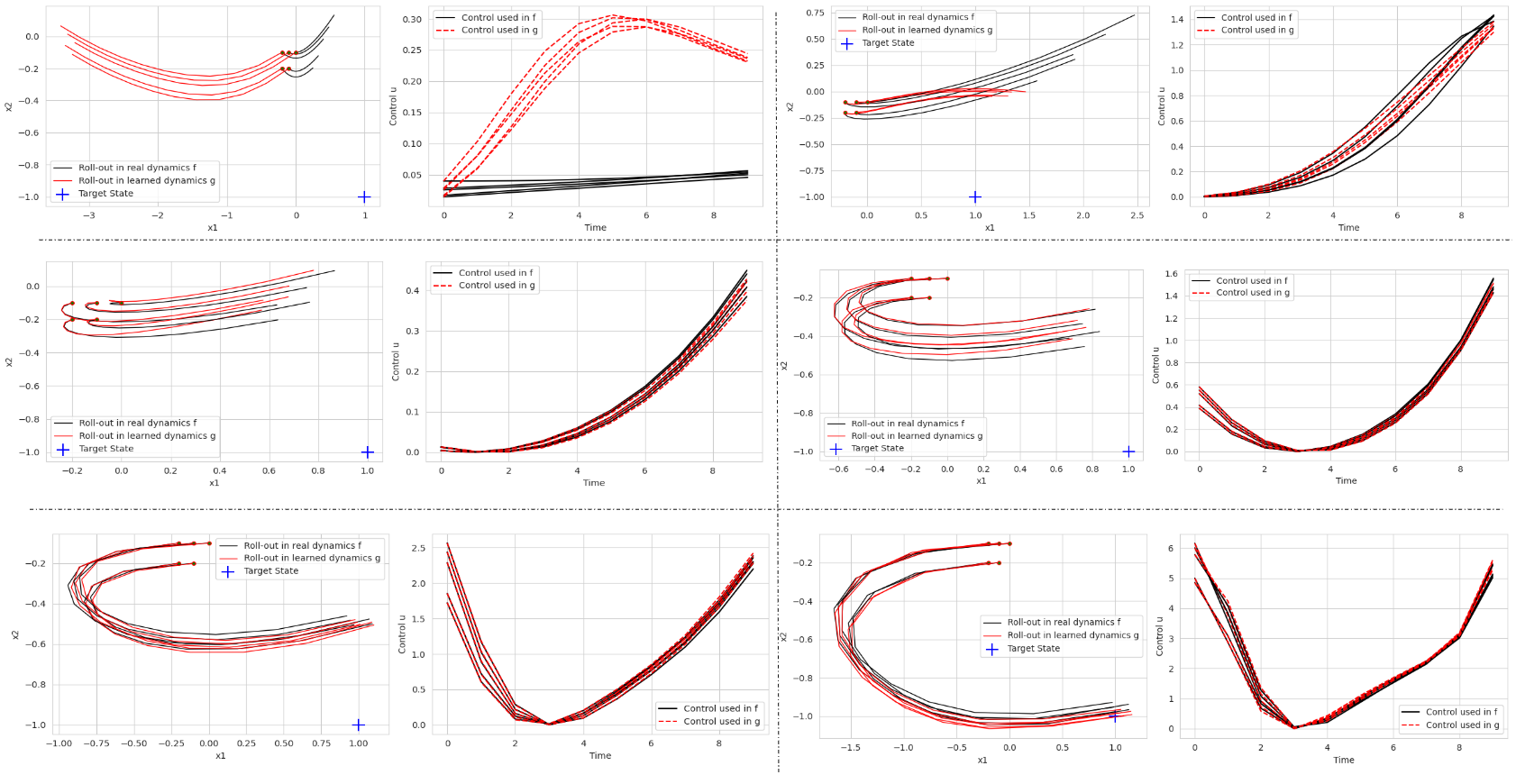}}
\caption{The full training process of NC for Ax+Bu task}. 
\label{fig:Full train}
\vskip -0.2in
\end{figure*} 

For such systems, analytical optimal control (OC) inputs can be derived using the Maximum Principle described in Equation (\ref{eq:max_principle}). 
The optimal control input $u^*(t)$ is given by the equation:
\begin{equation}
\label{eq:AxBu optimal u}
\begin{aligned}
u^*(t) = B^\top e^{A^\top (T-t)} W(T)^{-1} v(T)
\end{aligned}
\end{equation}
$v(T) = x(T) - e^{A(T-0)}x(0)$ represents the difference between the target state $x(T)$ and the initial state $x(0)$ under free evolution, as denoted in Equation (\ref{eq:AxBu optimal u}). The matrix $W(T)$ corresponds to the controllability Gramian, defined as:
\begin{equation}
\begin{aligned}
W(T) = \int_{0}^{T} e^{At} BB^\top e^{A^\top t} dt.
\end{aligned}
\end{equation}

This control task is the same task \cite{AI_Pontryagin} used, but we assume the dynamics is unknown, and our NC model $ {g}_{\gamma}(x, {h}_{\theta}(x, t), t)$ does not have access to the true dynamics $f(x,u)$. We aim to evaluate whether our NC can learn the analytical optimal control without the knowledge of the true dynamics. We set problem dynamics parameters: A = $\begin{pmatrix} 0  1 \\ 1  0 \end{pmatrix}$ and the driver matrix B = $\begin{pmatrix} 1 \\ 0 \end{pmatrix}$. We use a batch of uniformly random five initial states, the target state is set to be $x^* = (1,-1)$, and control horizon is set to be $T = 1$. Therefore, our control objective is to bring the system state $x_T$ at $T=1$ closer to the target state starting from those initial states $x_0$ in the randomly generated batch. 

\textbf{Algorithm set-up}

For solving this linear control task with our NC model $ {g}_{\gamma}(x, {h}_{\theta}(x, t), t)$, we employ a controller $ {h}_{\theta}$ and a dynamics learner $ {g}_{\gamma}$ with one hidden layer consisting of 30 neurons each. The output layer of both neural networks uses the identity activation function, while the hidden layer uses the tanh non-linear activation function.

To train our model using Algorithm \ref{alg:alternative_training}, we set the training alternation parameter to K = 10, indicating that it only needs to collect at most ten sets of real trajectories from the environment or the true dynamics. Within each alternation, we train both the dynamics learner and the controller for 10000 gradient steps ($n_g, n_c = 10000$), using a constant learning rate of 0.005. The training process leverages the built-in auto-differentiation functions in Jax. 

\textbf{Results}

Figure (\ref{fig:Full train}) demonstrate the full training process of NC when controlling the $Ax+Bu$ linear system. It provides an overview of the full alternative training process, consisting of six pairs of plots separated by dashed lines. Each pair corresponds to the results obtained in a specific training alternation $k$. The top-left pair corresponds to the initial training alternation ($k=0$), while the bottom-right pair represents the fully trained model after $K = 10$ alternations between the dynamics learner and the controller training. The pairs in between correspond to intermediate training alternations.

In each pair of plots, the left plot displays the trajectory roll-outs under the real dynamics $f$ (shown in black) and the current dynamics learner $ {g}_{\gamma}^k$ (shown in red), using the controller $h_{\theta}^k$ for the current alternation $k$. Meanwhile, the right plot shows the control signal generated along the trajectories using the current controller $h_{\theta}^k$. These control signals corresponds to the outputs of $h_{\theta}^k(x,t)$, where $x$ represents the states in the trajectories under $f$ and $ {g}_{\gamma}^k$, respectively.

The final trajectory plot at the bottom right corner reveals that the trained NC controller $h_{\theta}^K$ successfully guides the system from various initial states to the target state $x^* = (1,-1)$, in both learned dynamics and real dynamics (which shows that the training of the $g_{\gamma}^K$ is successful as well). The training losses for dynamics learner and controller corresponds to these six alternations (each alternation we take $n_c$, $n_g$ gradient steps) are shown in Figure (\ref{fig: train loss Ax+Bu}) Appendix \ref{app:training losses}. 

Moreover, it is fascinating to observe that the controller and dynamics learner converge almost simultaneously: as the controller performs better (end states in trajectories are closer to the target state), the dynamics learner adjusts to fit the dynamics more accurately (deviation between black and red trajectories are smaller). As discussed in Section \ref{sec:alternative training}, an intriguing interaction unfolds between the controller and the dynamics learner. The controller effectively guides the dynamics learner's learning process by leading it to different regions of the vector field in $f$ that require accurate fitting. Conversely, the dynamics learner imposes constraints on the controller by specifying the dynamics that the controller will operate within. In essence, they mutually shape each other's training objectives and supervise one another during each alternation $k$. As the alternations progress, the quality of this mutual supervision between the two neural networks continually improves. As the controller's performance enhances, the dynamics learner learns to approximate dynamics that are closer to optimal trajectories in the state space. Simultaneously, as the dynamics learner becomes more proficient, the controller's training becomes more effective in terms of the objective of steering the states closer to the target states, as the dynamics that the controller  operates in is closer to the true dynamics. This iterative feedback loop reinforces their collaborative training and leads to increasingly refined coordination between the controller and the dynamics learner.

These findings once again highlight the captivating interactions and relationships between the controller $h_{\theta}$ and dynamics learner $g_{\gamma}$ in our NC model $ {g}_{\gamma}(x, {h}_{\theta}(x, t), t)$. Through this enthralling interplay between the two neural networks, we can solve optimal control tasks without any prior knowledge of the system dynamics.

\subsection{Control Task: CartPole}
\label{sec:cartpole}
We now investigate the control performance of NC applied to a non-linear dynamical system known as CartPole or inverted pendulum. 

\textbf{Dynamical system and control objectives}

The CartPole system is a classic problem in control theory and reinforcement learning. It consists of a pole that is attached to a cart, which can move along a frictionless track.

\begin{figure}[ht]
\begin{center}
\centerline{\includegraphics[width=0.4\columnwidth]{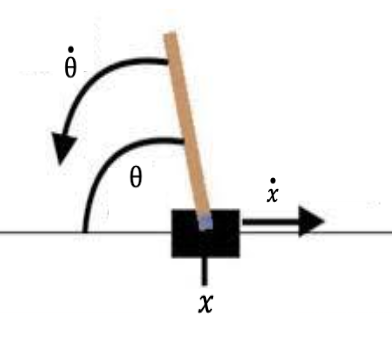}}
\caption{CartPole dynamical system}
\label{fig:cart_system}
\end{center}
\vskip -0.3in
\end{figure} 

The state of the system is typically represented by four variables: $(x, \dot x, \theta, \dot \theta)$ that represents the position of the pole, the horizontal velocity of the pole, the pole angle with respect to the vertical axis, and the pole's angular velocity, respectively. 

The control objective is to maintain the pole in an upright position (keep system state at (0,0,0,0)) by applying appropriate control inputs, which is the horizontal force applied to the cart. The CartPole system is a challenging problem because it is inherently unstable. Small changes in the control inputs or disturbances can cause the pole to fall. 

The dynamics of the system $f(x,\dot x, \theta, \dot \theta)$ that inputs the four-dimensional state and returns $\dot x, \ddot{x}, \dot \theta, \ddot{\theta}$ (the instantaneous change to the state) is given by the following system of differential equations \cite{CartPole_correct}: 
\begin{equation}
\label{eq:cartpole}
    \begin{aligned}
        \dot{x} = \dot x \\
        \ddot{x} = \frac{{F + m_p \ l \ (\dot{\theta}^2 \sin\theta - \Ddot{\theta} \cos\theta)}}{{m_c + m_p}} \\
        \dot{\theta} = \dot \theta \\
        \ddot{\theta} = \frac{{g\sin\theta + \cos\theta \left(\frac{{-F - m_p \ l \ \dot{\theta}^2 \sin\theta}}{{m_c + m_p}}\right)}}{{l \cdot \left(\frac{4}{3} - \frac{{m_p \cos^2\theta}}{{m_c + m_p}}\right)}}
    \end{aligned}
\end{equation}
Here, $F$ represents the control input applied to the cart, which is determined by the controller. The gravitational acceleration is $g=9.8$. The masses of the cart and the pole are represented by $m_c = 1.0$ and $m_p=0.1$ respectively. The length of the pole is $l = 0.5$ (half of the actual pole length). Equation (\ref{eq:cartpole}) shows that the system dynamics are highly non-linear. Note that the first line $\dot x = \dot x$ simply means that the first dimension of the output of $f$ (how $x$ should change) is directly equal to the third dimension of the input to $f$. 

\begin{figure*}[!t]
\centerline{\includegraphics[width=0.85\textwidth]{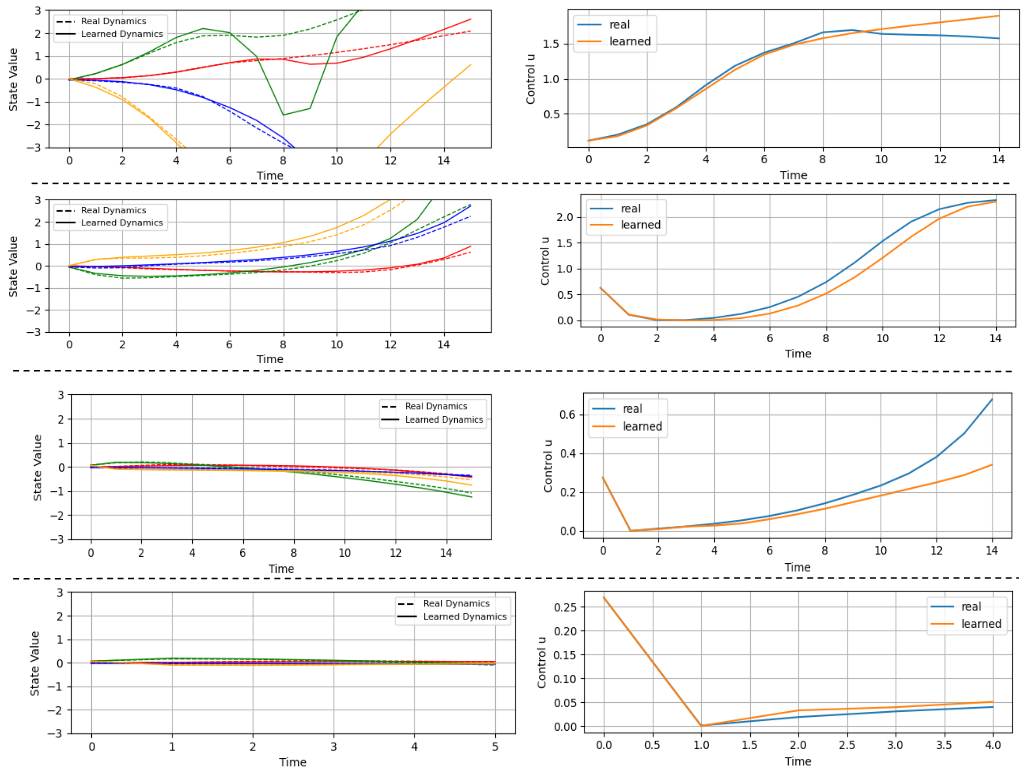}}
\caption{The full training process of NC for CartPole task}. 
\label{fig:Full train cart}
\vskip -0.2in
\end{figure*} 

An important observation is that the control loss in our NC model for the CartPole task has transitioned from a single-end-state loss to an integral loss. The control loss now measures the mean squared error between the desired state (0,0,0,0) and all states along the trajectories. This is because the objective is to maintain the pole in an upright position throughout the entire trajectory. This can be easily incorporated into our NC alternative training algorithm by changing line 17 of Algorithm \ref{alg:alternative_training}.

\textbf{Algorithm Set-up}

For solving the CartPole control task with our NC model $ {g}_{\gamma}(x, {h}_{\theta}(x, t), t)$, we set up our controller $ {h}_{\theta}$ with a single hidden layer consists of 64 neurons and one-dimensional output for control F, and our dynamics learner ${g}_{\gamma}$ with two hidden layers consisting of 64 and 32 neurons respectively and having a four-dimensional output corresponds to $\dot x, \ddot{x}, \dot \theta, \ddot{\theta}$. The output layer for both $ {h}_{\theta}$ and $ {g}_{\gamma}$ uses the identity activation function, while the hidden layer uses the tanh non-linear activation function. We set the training alternation parameter to $K = 12$ in Algorithm \ref{alg:alternative_training}. During each alternation, we train the controller for $n_c = 2000$ gradient steps with a learning rate of 0.0001. The training of the dynamic learner follows a multi-stage approach that enhances the stability of training. Firstly, we minimize the mean squared errors of trajectory roll-outs from $t=0$ to $t=0.5$ for 2000 gradient steps, using a learning rate of 0.005. Next, we extend the training to cover the time range from $t=0$ to $t=1$ for an additional 2000 gradient steps, with a learning rate of 0.001. Finally, we continue the training from $t=0$ to $t=T$ for another 2000 gradient steps, employing a learning rate of 0.0001. This multi-stage training can be seen in Figure (\ref{fig: train loss cart}) Appendix \ref{app:training losses}, where the dynamics learner loss plots have spikes for every 2000 gradient steps. 
On the other hand, the dynamic trainer follows a multi-stage training approach for the purpose of stability. AT first, the time step is set to $t = 0.5$ with $n_c = 10000$ gradient steps and a learning rate of 0.001. Then, the time step is extended to a time range of $t = 0.5$ to $t = 1$ with an additional 10000 gradient steps $n_c$ and a learning rate of 0.0001.

\textbf{Results}

Figure (\ref{fig:Full train cart}) indicates the full training process of NC for controlling the CartPole system with four pairs of plots. The pair on the top corresponds to the initial training alternation ($k=0$), while the bottom represents the fully trained model after $K = 12$ alternations. The pairs in the middle correspond to intermediate training alternations. In each plot pair, the left plot illustrates the trajectory roll-outs under the real dynamics $f$ (shown in dashed lines) and under the current dynamics learner $ {g}_{\gamma}^k$ using the controller. Since the states in this case have four dimensions, we represent the trajectory roll-outs by plotting the values of each state dimension over time, rather than directly visualizing the trajectories in the state space. It is important to note that we visualize the results for a single initial state from the batch of data. The right plot in each pair displays the control signals generated along the trajectories.

We can observe that NC controller $h_{\theta}^K$ successfully controls the CartPole system after K alternations, as it keeps the system evolution near the target state (0,0,0,0) which corresponds to the upright position of the pole, and the trajectory roll-outs in learned dynamics are close to real dynamics. Again, the training losses for dynamics learner and controller correspond to these four alternations are shown in Figure (\ref{fig: train loss cart}) Appendix \ref{app:training losses}. 

One thing to note here is the data efficiency of NC algorithm. As indicated in Algorithm \ref{alg:alternative_training} line 5, we only need to interact with the real environment and collect real trajectories at each alternation $k$ for each initial state in our batch. Therefore, we only need to collect in total $K*N$ trajectories or episodes from the real environment where $N$ is the batch size. For this particular experiment, we only collected less than 60 trajectories before successfully solving the CartPole tasks, which is more data-efficient compared to RL methods that uses hundreds or thousands of trajectories.

\section{Conclusion}
We introduced a novel framework denoted as NC that consists of a coupled neural ODE for controlling dynamical systems with unknown dynamics. This is achieved via an alternative training between two neural networks in our NC model: the dynamics learner and the controller. Through an intriguing interplay and mutual supervision between these two neural networks, the model learns the optimal control as well as the system dynamics. We applied NC in different dynamical systems and demonstrated the effectiveness of NC for learning the optimal control with high data efficiency. 

This manuscript presents a detailed and technical exposition of the method introduced in Chapter 3 ('TEL for Optimal Control: NC') of the thesis titled 'Theory Embedded Learning' available at \href{https://tspace.library.utoronto.ca/handle/1807/130447}{this link} \cite{chi2023theory}. 

%Bibliography
\bibliographystyle{unsrt}  
\bibliography{references} 

\begin{thebibliography}{10}

\bibitem{oc}
Wiley~Online Library.
\newblock Optimal control applications and methods.
\newblock {\em Wiley Online Library}, 2023.

\bibitem{gelfand1963calculus}
I.~M. Gelfand and S.~V. Fomin.
\newblock {\em Calculus of Variations}.
\newblock Prentice-Hall, 1963.

\bibitem{max_principle1}
Pontryagin L Boltyanskii V. Gamkrelidze~R. Mishchenko.
\newblock {\em Mathematical Theory of Optimal Processes}.
\newblock 1961.

\bibitem{max_principle2}
E.~McShane.
\newblock The calculus of variations from the beginning through optimal control theory.
\newblock {\em SIAM J. Control Optim}, 26:916--939, 1989.

\bibitem{bellman1957dynamic}
Richard Bellman.
\newblock {\em Dynamic Programming}.
\newblock Princeton University Press, 1957.

\bibitem{AI_Pontryagin}
Lucas B{\"o}ttcher, Nino Antulov-Fantulin, and Thomas Asikis.
\newblock Ai pontryagin or how artificial neural networks learn to control dynamical systems.
\newblock {\em Nature Communications}, 13(1):333, 2022.

\bibitem{first_NODEC}
Thomas Asikis, Lucas B\"ottcher, and Nino Antulov-Fantulin.
\newblock Neural ordinary differential equation control of dynamics on graphs.
\newblock {\em Phys. Rev. Res.}, 4:013221, Mar 2022.

\bibitem{system_identification_soderstrom1989system}
Torsten S{\"o}derstr{\"o}m and Petre Stoica.
\newblock {\em System Identification}.
\newblock Prentice Hall, 1989.

\bibitem{chen2022neural}
Tianju Chen, Yulia Rubanova, Jesse Bettencourt, and David Duvenaud.
\newblock Neural ordinary differential equations for nonlinear system identification, 2022.

\bibitem{author2023accurate}
Anonymous.
\newblock How accurate are neural approximations of complex network dynamics?, 2023.

\bibitem{holl2020learning}
Philipp Holl, Vladlen Koltun, and Nils Thuerey.
\newblock Learning to control pdes with differentiable physics, 2020.

\bibitem{TRPO}
John Schulman, Sergey Levine, Philipp Moritz, Michael~I Jordan, and Pieter Abbeel.
\newblock Trust region policy optimization.
\newblock {\em arXiv preprint arXiv:1502.05477}, 2015.

\bibitem{PG_RL}
Richard~S Sutton, David~A McAllester, Satinder~P Singh, and Yishay Mansour.
\newblock Policy gradient methods for reinforcement learning with function approximation.
\newblock {\em Advances in neural information processing systems}, 12(1):1057--1063, 2000.

\bibitem{AC_RL}
Vijay~R Konda and John~N Tsitsiklis.
\newblock Actor-critic algorithms.
\newblock {\em Advances in neural information processing systems}, 12(1):1008--1014, 2000.

\bibitem{model-based-RL}
Thomas~M. Moerland, Joost Broekens, Aske Plaat, and Catholijn~M. Jonker.
\newblock Model-based reinforcement learning: A survey.
\newblock 2020.

\bibitem{diffuser}
Michael Janner, Yilun Du, J.~Tenenbaum, and S.~Levine.
\newblock Planning with diffusion for flexible behavior synthesis.
\newblock In {\em International Conference on Machine Learning}, May 2022.
\newblock DOI: 10.48550/arXiv.2205.09991, Corpus ID: 248965046.

\bibitem{low_eff_RL}
Max Schwarzer, Ankesh Anand, Rishab Goel, R~Devon Hjelm, Aaron Courville, and Philip Bachman.
\newblock Data-efficient reinforcement learning with self-predictive representations.
\newblock In {\em International Conference on Learning Representations (ICLR)}, 2021.

\bibitem{rim}
Amine~Mohamed Aboussalah, Minjae Kwon, Raj~G Patel, Cheng Chi, and Chi-Guhn Lee.
\newblock Recursive time series data augmentation.
\newblock In {\em International Conference on Learning Representations (ICLR)}, February 2023.
\newblock Published: 01 Feb 2023, Last Modified: 13 Feb 2023.

\bibitem{neuralODE}
Ricky~TQ Chen, Yulia Rubanova, Jesse Bettencourt, and David Duvenaud.
\newblock Neural ordinary differential equations.
\newblock In {\em Advances in Neural Information Processing Systems}, pages 6571--6583, 2018.

\bibitem{Boettcher2022}
Lucas B{\"o}ttcher and Thomas Asikis.
\newblock Near-optimal control of dynamical systems with neural ordinary differential equations.
\newblock {\em Machine Learning: Science and Technology}, 3(4):045004, 2022.

\bibitem{neural_odes_feedback_2022}
Neural odes as feedback policies for nonlinear optimal control, 2022.

\bibitem{learning_dynamics_partial_2022}
Learning dynamics from partial observations with structured neural odes, 2022.

\bibitem{differential_neural_network_2023}
Differential neural network identifier for dynamical systems with time-varying state constraints, 2023.

\bibitem{nerualODE}
Ricky~TQ Chen, Yulia Rubanova, Jesse Bettencourt, and David Duvenaud.
\newblock Neural ordinary differential equations.
\newblock {\em Advances in neural information processing systems}, 31:6571--6583, 2018.

\bibitem{cartpole_RL}
Swagat Kumar.
\newblock Balancing a cartpole system with reinforcement learning, 2020.

\bibitem{example1}
Gang Yan, Jie Ren, Ying-Cheng Lai, Choy-Heng Lai, and Baowen Li.
\newblock Controlling complex networks: How much energy is needed?
\newblock {\em Phys. Rev. Lett.}, 108:218703, May 2012.

\bibitem{CartPole_correct}
Razvan~V. Florian.
\newblock Correct equations for the dynamics of the cart-pole system.
\newblock 2005.

\bibitem{chi2023theory}
Cheng Chi.
\newblock {\em Theory Embedded Learning}.
\newblock PhD thesis, Mechanical and Industrial Engineering, University of Toronto, Nov 2023.

\end{thebibliography}

\newpage
\appendix
\section{Alternative training motivation}
\label{app:motivation}

Figure \ref{fig:vector} below depicts the trajectory roll-outs under $ {g}{\gamma}(x, {h}{\theta}(x, t), t)$ using already well-trained $ {g}{\gamma}$ and $ {h}{\theta}$ (after K alternations in Figure \ref{fig:Full train}. The figure also includes visualizations of the true vector field of $f$ and the vector field of the learned dynamics $ {g}$. While the vector field approximation may not be ideal across most of the state space, it fits perfectly in the vicinity of the trajectory roll-outs (as indicated by the black circles where dark red and dark blue vector fields are aligned). This demonstrate the points we made in section \ref{sec:alternative training}: we do not need a perfect fit dynamics learner, but rather just a good local approximation of the dynamics near the state space where the current controllers can lead to. And this local good approximation already suffices for learning the optimal controls.

\begin{figure}[H] % Add the [H] specifier here
\begin{minipage}{\columnwidth}
\vskip 0.2in
\centering
\includegraphics[width=0.4\columnwidth]{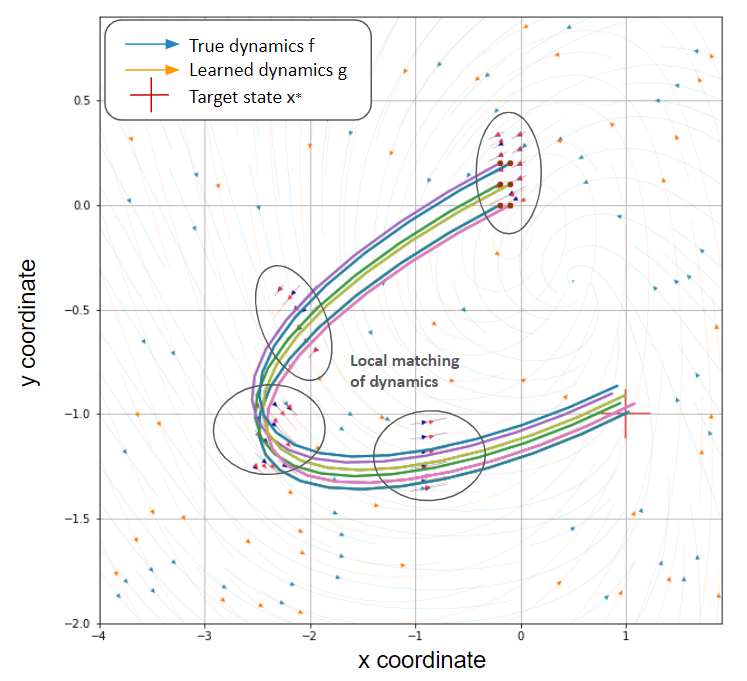}
\caption{Vector field visualization of trained NC for Ax + Bu task}
\label{fig:vector}
\end{minipage}
\end{figure}

\section{Compare with analytical optimal control}
\begin{figure}[H] % And here
\begin{minipage}{\columnwidth}
\vskip 0.2in
\centering
\includegraphics[width=0.6\columnwidth]{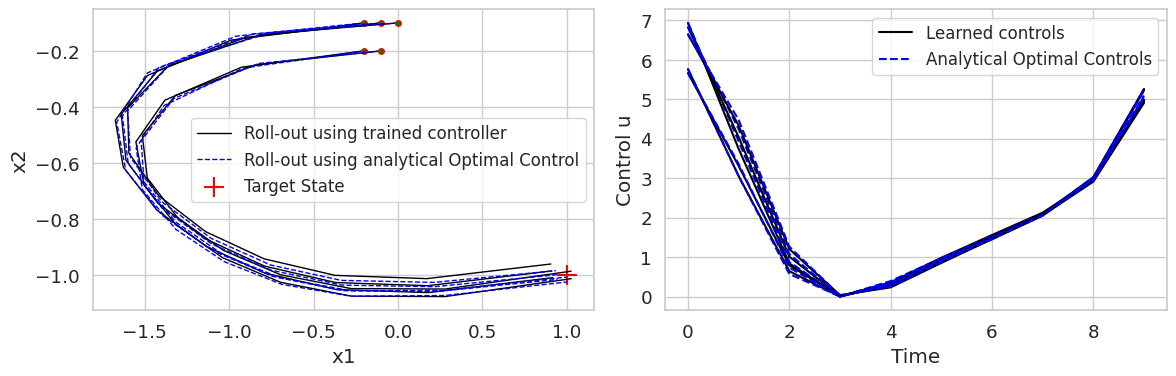}
\caption{Comparison of NC controller with optimal controller}
\label{fig:vsOC}
\end{minipage}
\end{figure}

Here we compare the trained NC controller in section \ref{Exp} Figure \ref{fig:Full train} with the analytical optimal controller given by Equation \ref{eq:AxBu optimal u}. 
The left plot in Figure \ref{fig:vsOC} above shows the trajectory roll-out in learned dynamics (which is almost the same as the real dynamics as indicated by the last pair of plots in Figure \ref{fig:Full train}). The black trajectories are roll-out using trained controller with NC (the same trajectories in the last pair in Figure \ref{fig:Full train}), while the blue trajectories are roll-out using optimal controllers. The high similarity between trajectory roll-outs as well as control signals indicate that through alternative training, our NC model can indeed learn optimal controls without any prior knowledge about system dynamics. 

%%%%%%%%%%%%%%%%%%%%%%%%%%%%%%%%%%%%%%%%%%%%%%%%%%%%%%%%%%%%%%%%%%%%%%%%%%%%%%%
%%%%%%%%%%%%%%%%%%%%%%%%%%%%%%%%%%%%%%%%%%%%%%%%%%%%%%%%%%%%%%%%%%%%%%%%%%%%%%%
\section{Dynamic learner and Controller training loss}
\label{app:training losses}
\begin{figure*}[ht]
\centerline{\includegraphics[width=0.9\textwidth]{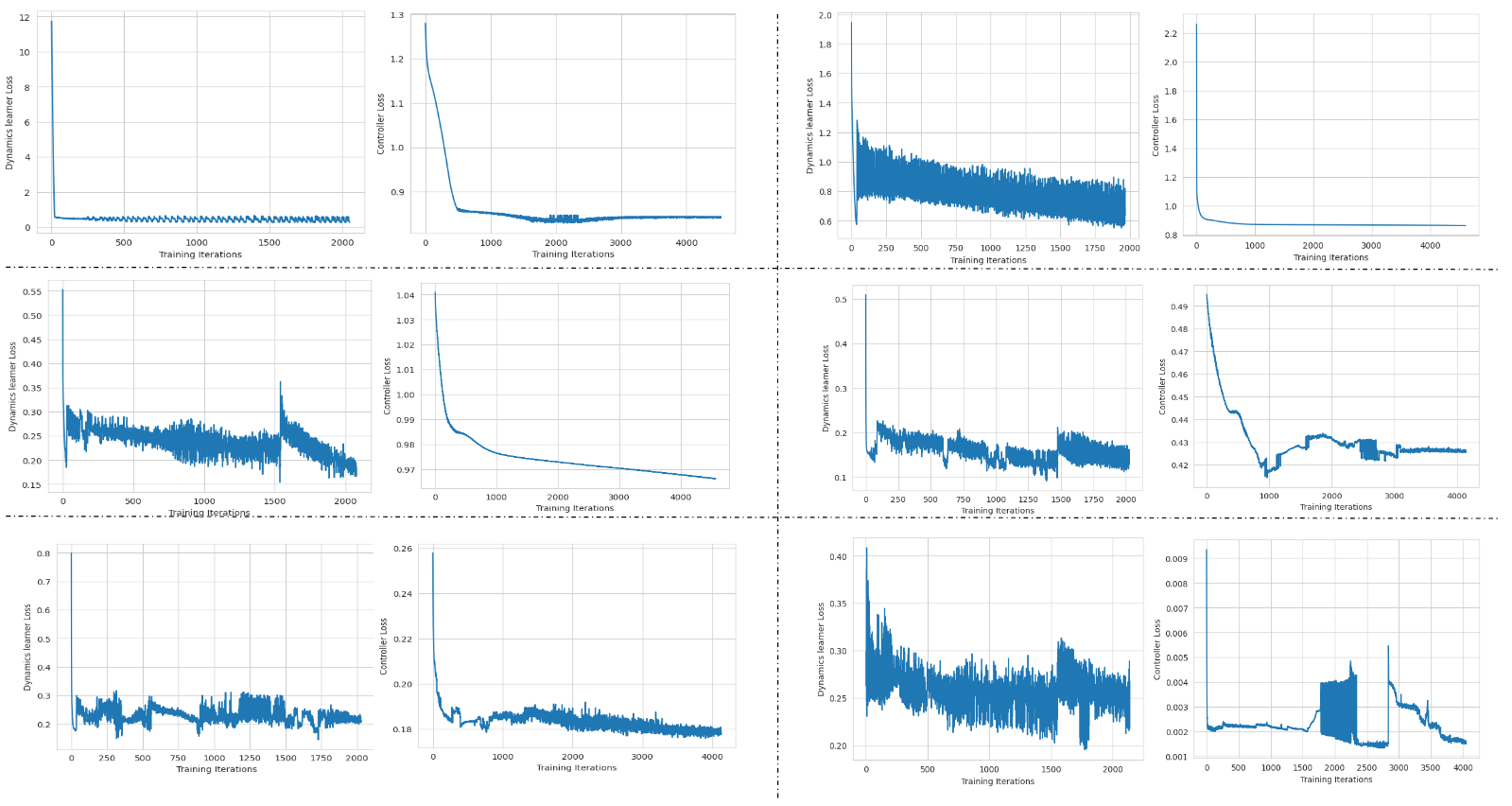}}
\caption{Dynamics learner training loss and controller training loss over K alternations for \textbf{Ax + Bu} task}
\label{fig: train loss Ax+Bu}
\end{figure*} 

Please note that Figure (\ref{fig: train loss Ax+Bu}) contains a total of 6 plot pairs, corresponding to the same alternations depicted in Figure (\ref{fig:Full train}) with the exact layout. 

\label{app:training losses}
\begin{figure*}[ht]
\centerline{\includegraphics[width=0.9\textwidth]{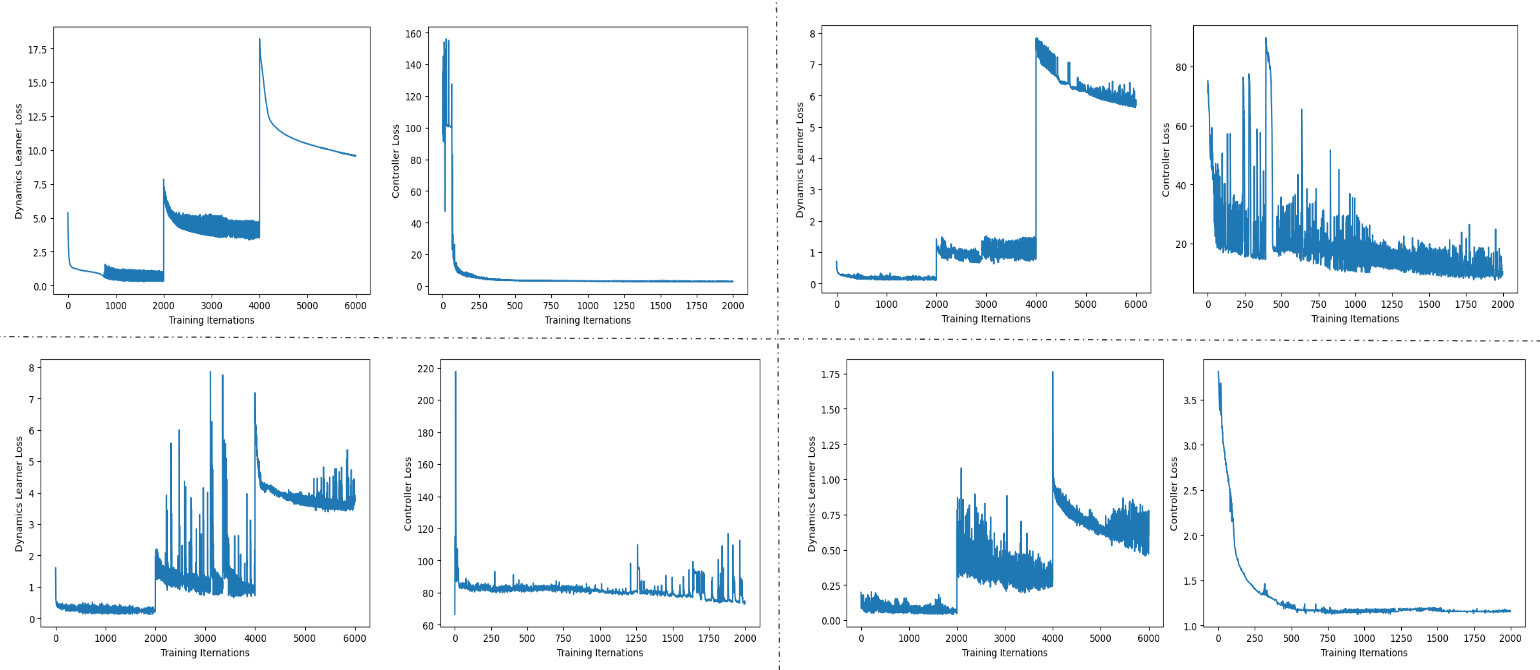}}
\caption{Dynamics learner training loss and controller training loss over K alternations for \textbf{CartPole} task}
\label{fig: train loss cart}
\end{figure*} 
Please note that Figure (\ref{fig: train loss cart}) contains a total of 4 plot pairs, corresponding to the same alternations depicted in Figure (\ref{fig:Full train cart}). The sequence of the plot pairs is as follows: top left, top right, bottom left, bottom right.

\end{document}